\documentclass[letterpaper]{article}
\usepackage{aaai}
\usepackage{times}
\usepackage{helvet}
\usepackage{courier}
\usepackage{hyperref}
\usepackage[utf8]{inputenc} 
\usepackage[T1]{fontenc}    
\usepackage{hyperref}       
\usepackage{url}            
\usepackage{booktabs}       
\usepackage{amsfonts}       
\usepackage{nicefrac}       
\usepackage{microtype}      
\usepackage{xcolor}         

\usepackage{amssymb}
\usepackage{amsmath}
\usepackage{mathabx}
\usepackage{stmaryrd}
\usepackage{graphicx}
\usepackage{subcaption}
\usepackage[ruled,vlined,noend]{algorithm2e}

\newtheorem{remark}{Remark}

\newtheorem{conjecture}{Conjecture}

\begin{document}
%
\title{GPINN: Physics-informed Neural Network with Graph Embedding}
\author{Yuyang Miao$^1$, Haolin LI$^{2}$\thanks{Corresponding author.}\\
$^1$ Department of Electrical and Electronic Engineering, Imperial College London\\
London SW7 2AZ, United Kingdom\\
$^2$ Department of Aeronautics, Imperial College London\\
London SW7 2AZ, United Kingdom\\
\{yuyang.miao20, haolin.li20\}@imperial.ac.uk
}
\maketitle
\begin{abstract}
\begin{quote}
This work proposes a Physics-informed Neural Network framework with Graph Embedding (GPINN) to perform PINN in graph, i.e. topological space instead of traditional Euclidean space,  for improved problem-solving efficiency. The method integrates topological data into the neural network's computations, which significantly boosts the performance of the Physics-Informed Neural Network (PINN). The graph embedding technique infuses extra dimensions into the input space to encapsulate the spatial characteristics of a graph while preserving the properties of the original space. The selection of these extra dimensions is guided by the Fiedler vector, offering an optimised pathologic notation of the graph. Two case studies are conducted, which demonstrate significant improvement in the performance of GPINN in comparison to traditional PINN, particularly in its superior ability to capture physical features of the solution.
\end{quote}
\end{abstract}

\section{Introduction}
\label{sec:s1}
The Physics-Informed Neural Network (PINN) has demonstrated potential in solving partial differential equations (PDEs), generating solutions embodied within the framework of neural networks \cite{raissi2019physics}. This differs from conventional numerical approaches, such as the Finite Element Method (FEM), which utilize the weak form of PDEs to obtain discrete solutions. The PINN paradigm, however, uniquely leverages the strong form of PDEs, thereby yielding solution expressions that are continuous and differentiable \cite{raissi2019physics}. The robustness of PINNs has been underscored by their capacity to address inverse problems. Such problems often prove unassailable for FEM due to absence of complete boundary conditions in certain complex scenarios \cite{chen2020physics}.

A remarkable contribution of the Physics-informed Neural Network (PINN) approach is to integrate physical principles into the equation-solving process. This is achieved by calculating the residual of a partial differential equation (PDE) in its strong form, given that the neural network (NN) solution is differentiable throughout the entire effective domain. This residual is subsequently included in the loss function during NN training \cite{raissi2019physics}. The incorporation of this physical information substantially enhances the solution's robustness, mitigating the overfitting issue in forward problems and enabling the inference of global solutions from sparse local information in inverse problems \cite{grossmann2023can,chen2020physics}.

However, in traditional PINN model, the training of NN is only informed with the PDE information but not the space, i.e. the domain information at all, which is also pivotal in PDE problems. The reason is that the input space of PINN in traditional Euclidean space doesn't align consistently with the physical space: the Euclidean distance between two points may not be valid due to the bounded nature of the domain. This discrepancy poses significant challenges when employing PINN for problems associated with complex geometries or highly discontinuous solution fields, e.g. crack or fracture problems \cite{goswami2020transfer}. Therefore, enriching the PINN model with domain topology information and aligning the input space more closely with the physical properties could significantly enhance the PINN's performance.

In response to these challenges, we propose a Physics-informed Neural Network with Graph Embedding (GPINN) method. This novel approach incorporates extra dimensions to address the problems observed in higher dimensional topological space. These extra dimensions are informed by the graph theory, which quantitatively uncovers the influential relationships between different parts of the domain. The topological space acts as a prior, derived purely from the domain topology, and are integrated with the original Euclidean coordinates in the solving process.

The rest of this paper is organized as follows: Section 2 introduces GPINN, covers the basics of graph theory, and explains the method of determining extra dimensions using the Fiedler vector. Section 3 presents two case studies that apply the developed GPINN to a heat propagation problem and a cracking modeling method in solid mechanics, respectively. Finally, Section 4 offers some concluding remarks.

\section{Methodology} \label{sec:s2}
\begin{figure*}[t]
\centering
\includegraphics[width=0.75\textwidth]{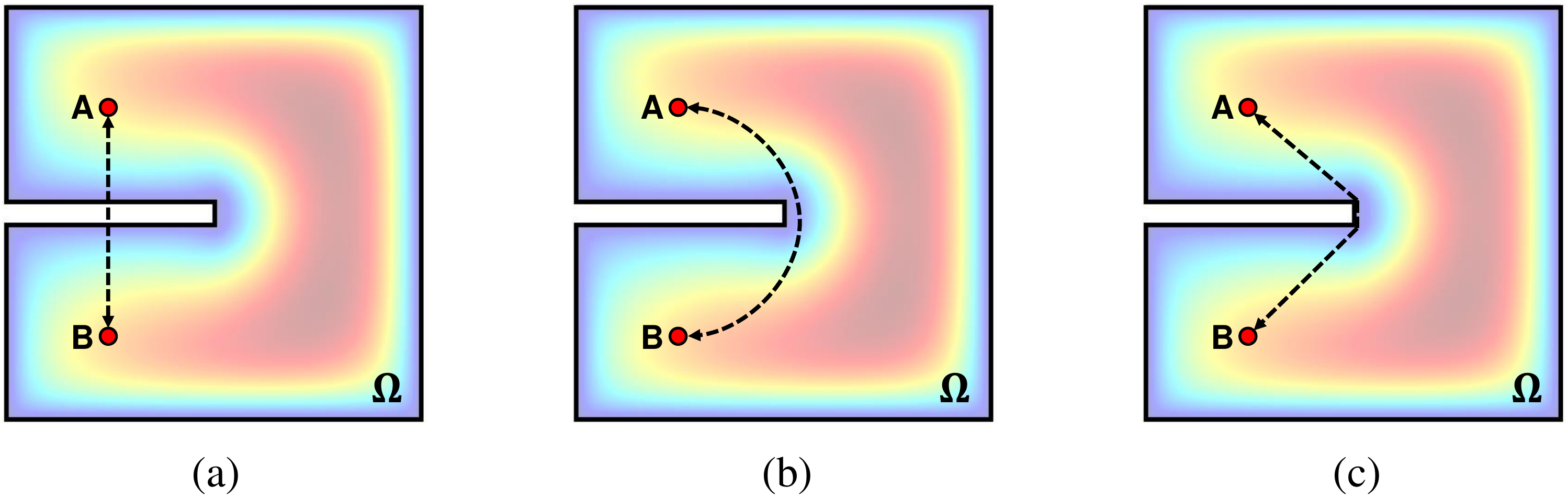}
\caption{Distances of heat propagation (a) in the input space, (b) in a possible propagation path in physics and (c) in the shortest propagation path in physics. }
\label{Fig_1}
\end{figure*}
This section presents the methods and fundamental knowledge underpinning our work. We first introduce the proposed Graph-Embedded Physics-Informed Neural Network (GPINN), followed by an overview of graph theory. We then discuss the application of graph theory in determining the extra dimension of GPINN.
\subsection{PINN with Graph Embedding} \label{sec:s2_1}
Physics-informed Neural Network dedicates to obtain data-driven solutions of partial differential equations (PDE) which generally takes the form \cite{raissi2019physics}:
\begin{equation}
\boldsymbol{u}(\boldsymbol{x},t) + \mathcal{N}(\boldsymbol{u}(\boldsymbol{x},t))=0, \boldsymbol{x} \in \Omega, t \in [0, T]
\label{e1}
\end{equation}
where $\boldsymbol{u}(\boldsymbol{x},t)$ represents a solution field dependent to the spatial ($\boldsymbol{x}$) and time ($t$) coordinates. The complexity of a physics system following such a partial differential rule usually makes it impossible to find a analytical solution of $\boldsymbol{u}(\boldsymbol{x},t)$. In this case, numerical methods are usually employed to obtain a approximate, typically discrete solution by its week form \cite{zienkiewicz2000finite}. However, PINN directly solve the strong form presented by Eq.\ref{e1} by expressing the solution by a neural network mapping that relates the inputs $x \in \Omega \subset \mathbb{R}^d$ and $t \in[0, T]$ to the output $u_(x,t) \in \mathbb{R}$ \cite{raissi2019physics}.
\begin{equation}
\boldsymbol{u}_{NN}(\boldsymbol{x},t): \Omega \rightarrow \mathbb{R}
\label{e2}
\end{equation}
Solving a partial differential equation is converted to a optimisation problem in PINN, in which the training of the developed neural network is required where the loss function is defined as:
\begin{equation}
\mathcal{L} = \omega_1 \mathcal{L}_{\mathrm{PDE}}+\omega_2 \mathcal{L}_{\mathrm{Data}}+\omega_3 \mathcal{L}_{\mathrm{IC}}+\omega_4 \mathcal{L}_{\mathrm{BC}}
\label{e3}
\end{equation}
where $\mathcal{L}_{\mathrm{PDE}}$, $\mathcal{L}_{\mathrm{Data}}$, $\mathcal{L}_{\mathrm{IC}}$ and $\mathcal{L}_{\mathrm{BC}}$ denote the residual of PDE i.e. $\mathcal{L}_{\mathrm{PDE}} = u(x,t) + \mathcal{N}(u(x,t))$, the loss of the sampling points, the loss of the initial condition and the loss of the boundary conditions, respectively, and $\omega$s are their scaling factors.

\begin{figure*}[h]
\centering
\includegraphics[width=0.80\textwidth]{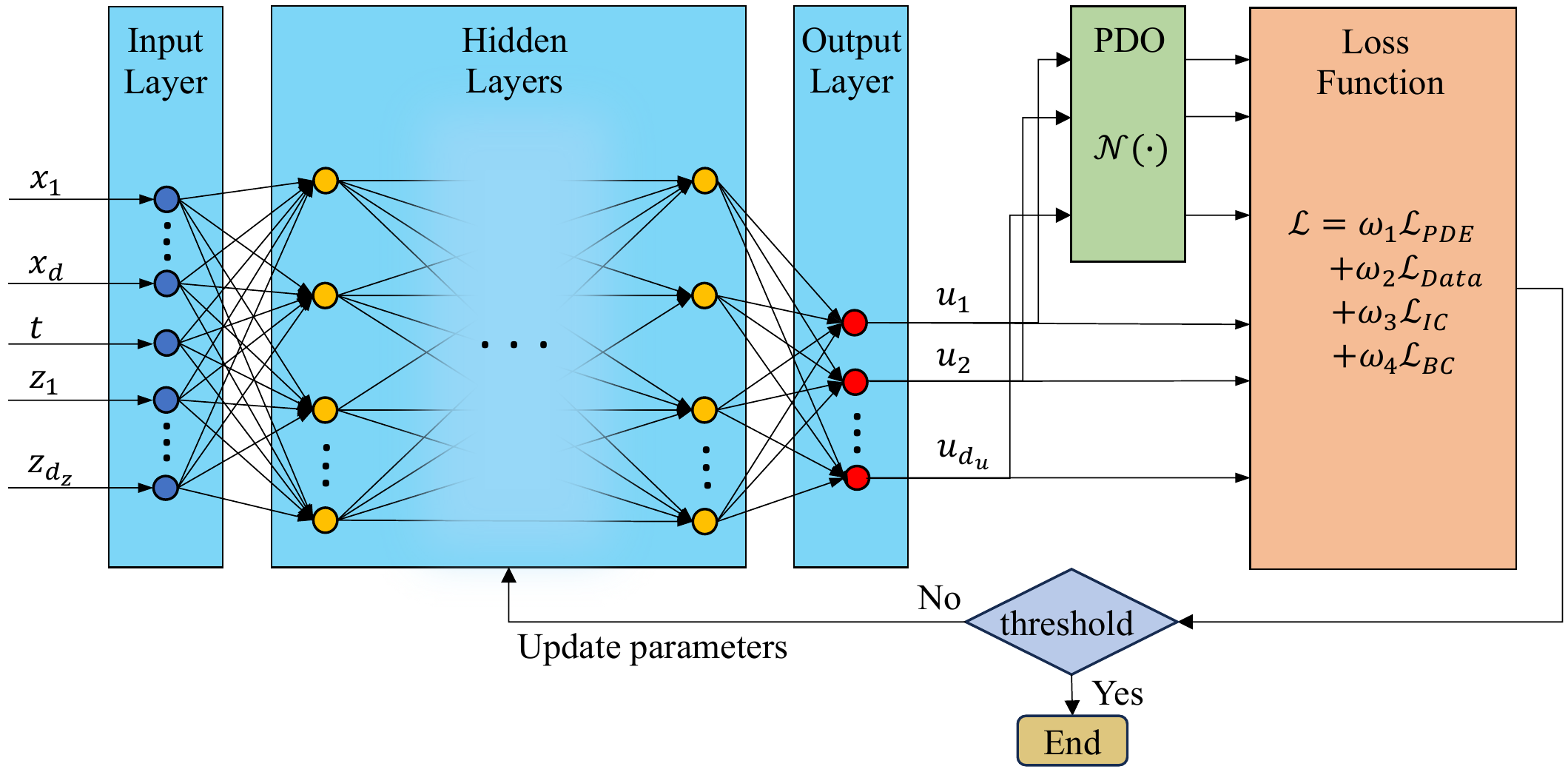}
\caption{Schematic of Physics-informed Neural Network with Graph Embedding. The neural network constructs the relation between the input spatial coordinate $\boldsymbol{x}$, time $t$ and extra dimensions $\boldsymbol{z}$ and the output $\boldsymbol{u}$.}
\label{Fig_2}
\end{figure*}

The data-driven solution offered by PINN is continuous and differentiable across the domain of $\mathbb{R}^d$ and the time interval $[0, T]$. This type of solution has distinct advantages over numerical solutions with condensed information. However, it also restricts its use in handling discontinuous fields or fields with non-differentiable segments. Furthermore, while the PINN solution provides a mapping between the input and solution spaces, the model itself doesn't strictly adhere to the physical essence. That is, the Euclidean distance between two points in the input space doesn't always correspond to their physical distance in $\Omega$.
An example of this discrepancy in distances is illustrated in Fig.\ref{Fig_1}, using a heat propagation scenario in a two-dimensional 'house' $\Omega$. The Euclidean distance between points A and B in PINN's input space is depicted in Fig.\ref{Fig_1}(a). However, in a physical context, heat actually propagates along the path shown in Fig.\ref{Fig_1}(b) or (c) for which the prescribed domain $\Omega$ prevents heat from travelling in a straight line.

To address this issue, this paper proposes a method that combines Graph Embedding into the Physics-informed Neural Network (PINN) to align the input space more closely with the physical attributes. As demonstrated by \cite{chung2011hot}, a topological space defined by graph theory more accurately captures physics-consistent characteristics compared to a Euclidean space. To this end, our approach transforms the operating space of PINN from the conventional Euclidean space to an approximated topological space by incorporating extra dimensions into the input space. Consequently, the solution of Eq.\ref{e1} is modified as follows:
\begin{equation}
\boldsymbol{u}_{NN}(\boldsymbol{x},t,\boldsymbol{z}): \Omega \rightarrow \mathbb{R}
\label{e4}
\end{equation}
in which the introduced extra dimensions are denoted by $\boldsymbol{z}$ and $\boldsymbol{z} \in \mathbb{R}^{d_z}$ where $d_z$ is the number of extra dimensions.
\begin{remark}
The GPINN methodology transforms the operating space from a Euclidean framework to a topological (graph-based) one through the incorporation of extra dimensions. This ensures a closer alignment between the problem domain and the physical attributes of the system under consideration.
\end{remark}
\begin{remark}
In this method, there is no need for additional coordinate correction in the original space, since the graph information is incorporated by extra dimensions in the PINN model that leaves the original inputs and their partial derivatives unaffected.
\end{remark}
\begin{remark}
The extra dimensions are uniquely defined by the specific geometry being studied, indicating that their identification is topology-specific and independent to the initial or boundary conditions.
\end{remark}
The subsequent subsections provide detailed information on determining extra dimensions for a prescribed domain. The architecture of the Physics-informed Neural Network with Graph Embedding is depicted in Fig.\ref{Fig_2}.

\subsection{Graph Theory} \label{sec:s2_2}
\begin{figure}[t]
    \centering
    \includegraphics[width=0.3\textwidth]{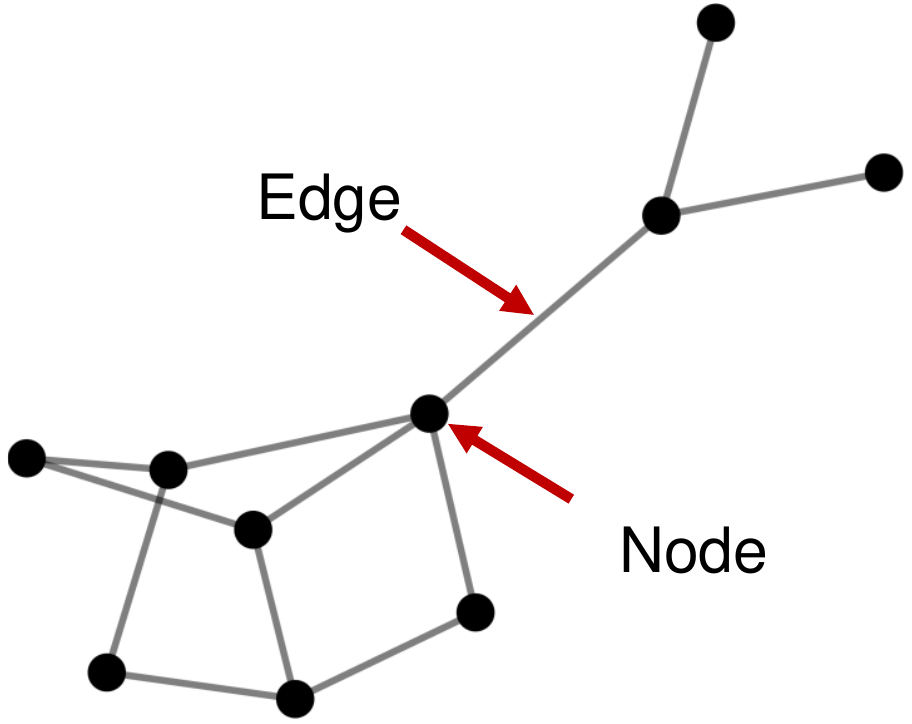}
    \caption{A example graph.}
    \label{fig:example_graph}
\end{figure}

A graph, denoted as $G = (V,E)$, is formed by a set of vertices $V$ interconnected by a set of edges $E$. As depicted in Fig. \ref{fig:example_graph}, vertices are symbolized as dots, and the lines that join them constitute the edges of the graph. The adjacency matrix $A$ denotes the connectivity details of a graph, with $A_{i,j} \neq 0$ indicating an edge between the $i^{th}$ and $j^{th}$ vertices.
Graphs can be categorized into two main types depending on whether their edges bear directionality: undirected and directed graphs. In the case of undirected graphs, the adjacency matrix is symmetric, thereby satisfying $A_{i,j} = A_{j,i}$.

This study exclusively considers undirected graphs due to the fact that the physical interactions are mutual. The degree of a vertex, a metric representing the number of nodes it connects to, can be found in the degree matrix $D$, where $D_{i,i}$ corresponds to the degree of vertex $i$. The graph Laplacian matrix is subsequently defined as $L = D - A$.

\subsection{Complex Geometries defined in Topological Space} \label{sec:s2_3}
A component's mesh can be conceptualized as a graph $G_{m}(V,E)$, where the set of vertices $V$ signifies the element points, and $E$ represents the edges. It's worth noting that a graph merely encodes the connectivity among nodes without preserving their exact positions. Intriguing insights can be gleaned when the mesh is treated strictly as a graph.

\begin{figure}[h]
    \centering
    \begin{subfigure}{0.5\textwidth}
        \centering
        \includegraphics[width=0.5\textwidth]{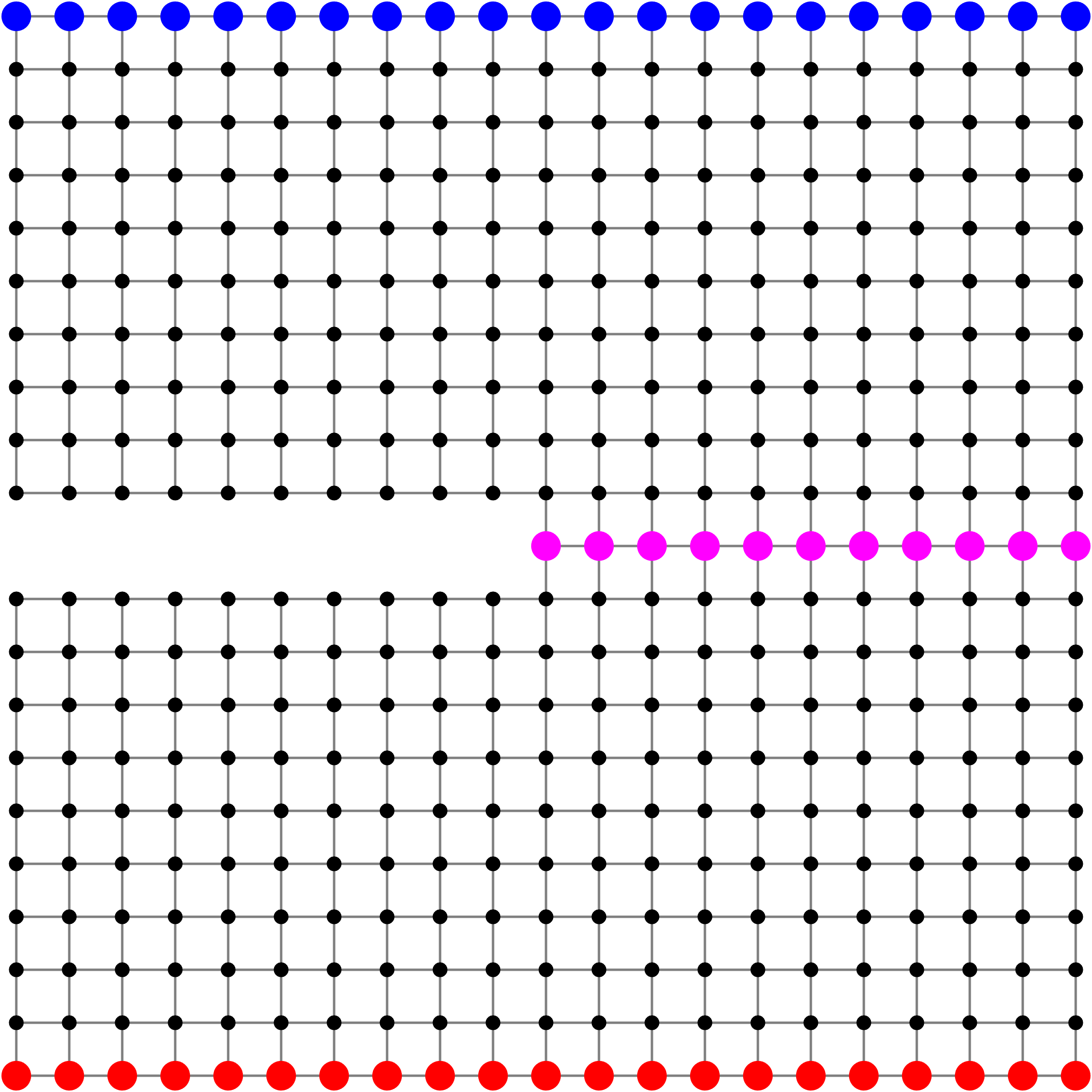}
        \caption{}
        \label{fig:Complex_Geometry}
    \end{subfigure}
    \hfill
    \begin{subfigure}{0.5\textwidth}
        \centering
        \includegraphics[width=0.7\textwidth]{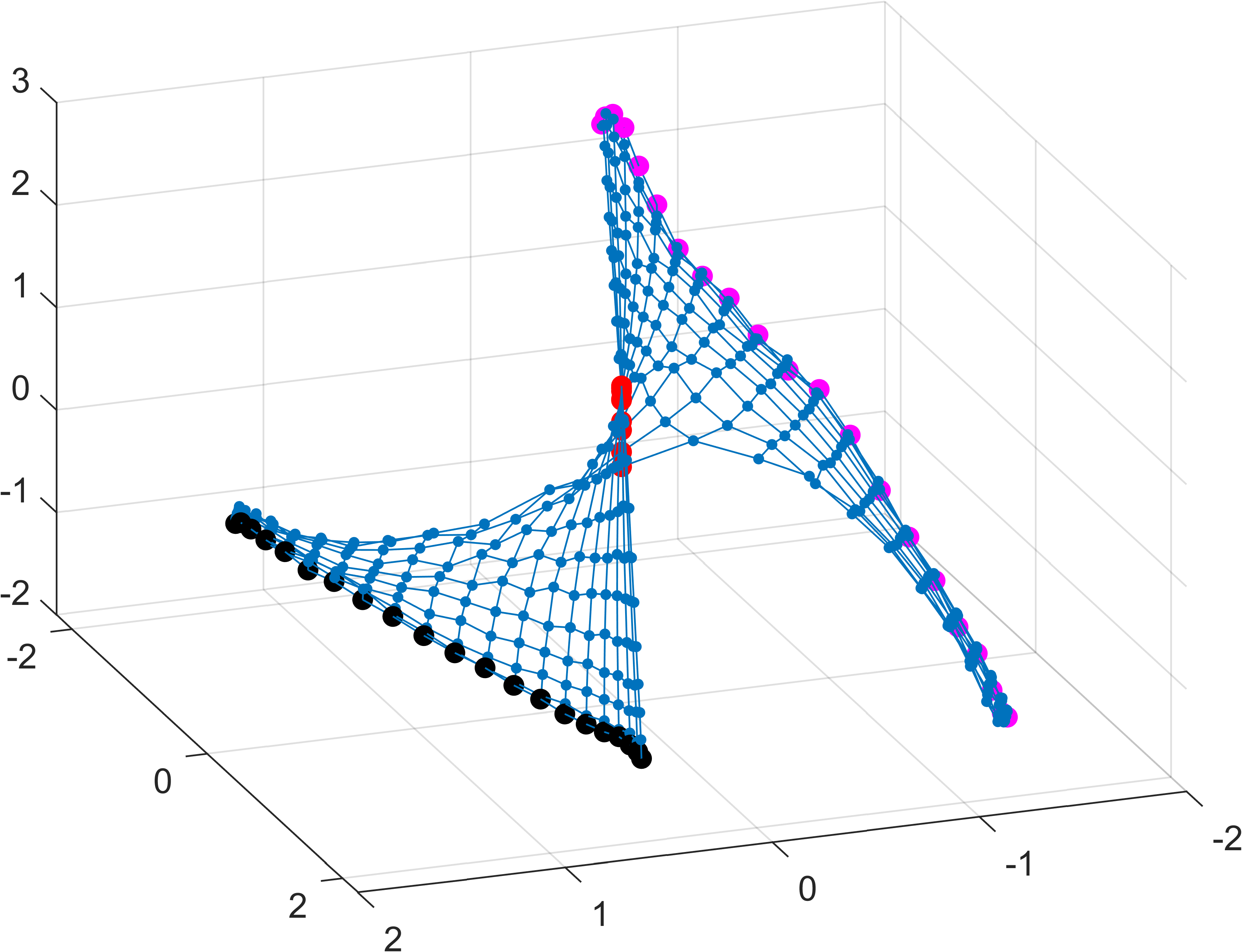}
        \caption{}
        \label{fig:Eigenspace}
    \end{subfigure}
    \caption{(a) Graph of a complex domain, (b) a possible graph layout in the eigenspace }
    \label{fig:whole}
\end{figure}

When the mesh is interpreted as a graph, it becomes apparent that many components with complex geometry resemble dumbbell-shaped graphs featuring dense clusters interconnected by tubes. An instance of such a connection is portrayed in Fig.\ref{fig:whole} for the structure presented in Fig.\ref{Fig_1}. Fig.\ref{fig:Eigenspace} illustrates the corresponding graph in the eigenspace . The component visualised in the Cartesian coordinates is depicted in Fig.\ref{fig:Complex_Geometry}. In both figures, nodes located at crucial positions are color-coded for enhanced comprehension.
In light of this perspective, the concept of dimension adaptive can be reconceived within the graph domain as graph labeling, in which the labels should:
\begin{itemize}
    \item Conform to the pattern of the component's overall shape.
    \item Maintain smoothness between clusters to simulate a continuous field on a graph.
\end{itemize}
The Fiedler vector potentially fulfills the above stipulations. It is the eigenvector of the graph Laplacian associated with the smallest non-zero eigenvalue. The subsequent chapter will establish why the Fiedler vector meets these criteria.

\subsection{Fiedler Vector: Relation with Heat Equation} \label{sec:s2_4}
Chun \textit{et al.} \cite{chung2011hot} attempted to show the relationship between the Fiedler vector and the discrete heat transfer equation. Consider a discrete heat transfer equation on a graph:
\begin{equation}
\frac{d \mathbf{f}}{d t}=\mathbf{L} \mathbf{f}, \mathbf{f}(0)=\mathbf{f}_{0}
\label{eq: Heat Function}
\end{equation}
where $\mathbf{L}$ denotes the Laplacian. The solution to Eq.\ref{eq: Heat Function} is:
\begin{equation}
\mathbf{f}(t)=\sum_{i=1}^{n}\left(\mathbf{f}_{0}, \mathbf{u}_{i}\right) e^{-\lambda_{i} t} \mathbf{u}_{i}
\label{eq: solution}
\end{equation}
where $e_{i}$ denotes the orthonormal basis of eigenvectors of L. Since $\lambda_{1} =0$ and $u_{1} =  [1,1,...,1,1]/\sqrt{N}$. The solution Eq.\ref{eq: solution} can be rewritten as:
\begin{equation}
\mathbf{f}(t) \approx\left(\mathbf{f}_{0}, \mathbf{u}_{1}\right) \mathbf{u}_{1}+\left(\mathbf{f}_{0}, \mathbf{u}_{2}\right) e^{-\lambda_{2} t} \mathbf{u}_{2} + R
\label{eq: rewrite}
\end{equation}
where the first term is the final state average signal and the remainder $R$ goes to zero faster than the term $\left(\mathbf{f}_{0}, \mathbf{u}_{2}\right) e^{-\lambda_{2} t} \mathbf{u}_{2} $. Therefore, the second eigenvector $u_{2}$ with a constant bias could model the transient state. Then recall the hot spot conjecture:
Conjecture 1. (Rauch's hot spots conjecture) Let \(\mathcal{M}\) be an open connected
bounded subset. Let \(f(\sigma, p)\) be the solution of heat equation, then
\begin{equation}
\frac{\partial f}{\partial \sigma}=\Delta f
\label{ef1}
\end{equation}
with the initial condition \(f(0, p)=g(p)\) and the Neumann boundary condition
\(\frac{\partial f}{\partial n}(\sigma, p)=0\) on the boundary \(\partial \mathcal{M}\). Then for most initial conditions, if \(p_{\text {hot }}\) is
a point at which the function \(f(\cdot, p)\) attains its maximum (hot spot), then the
distance from \(p_{\text {hot }}\) to \(\partial \mathcal{M}\) tends to zero as \(\sigma \rightarrow \infty\) \cite{banuelos1999hot}.
The minimum point (cold spot) follows a similar rule. Since the second eigenvector models the heat transient stage, we can employ the conjecture \cite{chung2011hot}:
\begin{conjecture}
Given a graph \(G=(V, E)\), If \(v^{*}, w^{*} \in V\), then
$$
\left|u_{2}(v)-u_{2}(w)\right| \leq\left|u_{2}\left(v^{*}\right)-u_{2}\left(w^{*}\right)\right| \quad \forall(v, w) \in V^{2}
$$
$$
d(v, w) \leq d\left(v^{*}, w^{*}\right) \quad \forall(v, w) \in V^{2}
$$
where \(d(v, w)\) is the geodesic between \(v\) and \(w\), in other words, the level of the extremeness of the Fiedler vector's value at a point reflects its geometric information.
\end{conjecture}
This hypothesis posits that the hot and cold spots present the greatest geodesic distance in comparison to any other pair of nodes. It also suggests that nodes positioned at a significant distance from each other will exhibit distinct values. Specifically, if two nodes are part of two separate dense clusters, their values will diverge due to the considerable geodesic distance across the tube.

The Fiedler Vector can be interpreted as a 1D embedding of the graph, where the values encapsulate information about the graph's structure. In relation to the requirement of smoothness, we can rephrase the definition of eigenvector and eigenvalue, $\mathbf{L}\boldsymbol{x} = \lambda \boldsymbol{x}$, as follows:
\begin{equation}
\begin{aligned}
\mathbf{u}_k^T \mathbf{L} \mathbf{u}_k & =\sum_{m=0}^{N-1} u_k(m) \sum_{n=0}^{N-1} A_{m n}\left(u_k(m)-u_k(n)\right) \\ & =\sum_{m=0}^{N-1} \sum_{n=0}^{N-1} A_{m n}\left(u_k^2(m)-u_k(m) u_k(n)\right)
\end{aligned}
\end{equation}
And owing to the symmetry of the adjacency matrix A ($A_{m,n} = A_{n,m}$):
\begin{equation}
\begin{aligned}
\mathbf{u}_k^T \mathbf{L} \mathbf{u}_k= & \frac{1}{2} \sum_{m=0}^{N-1} \sum_{n=0}^{N-1} A_{m n}\left(u_k^2(m)-u_k(m) u_k(n)\right) \\
& +\frac{1}{2} \sum_{m=0}^{N-1} \sum_{n=0}^{N-1} A_{m n}\left(u_k^2(n)-u_k(n) u_k(m)\right) \\
= & \frac{1}{2} \sum_{m=0}^{N-1} \sum_{n=0}^{N-1} A_{m n}\left(u_k(n)-u_k(m)\right)^2 = \lambda
\end{aligned}
\end{equation}
It is evident that the eigenvalues mirror the variation between nodes and their adjacent counterparts. Hence, the lower the eigenvalue, the smoother the corresponding eigenvector will be. Therefore, the second eigenvector will be the smoothest one, barring the first eigenvector, which is a constant vector.

\begin{figure}[h]
    \centering
    \begin{subfigure}{0.5\textwidth}
        \centering
        \includegraphics[width=0.7\textwidth]{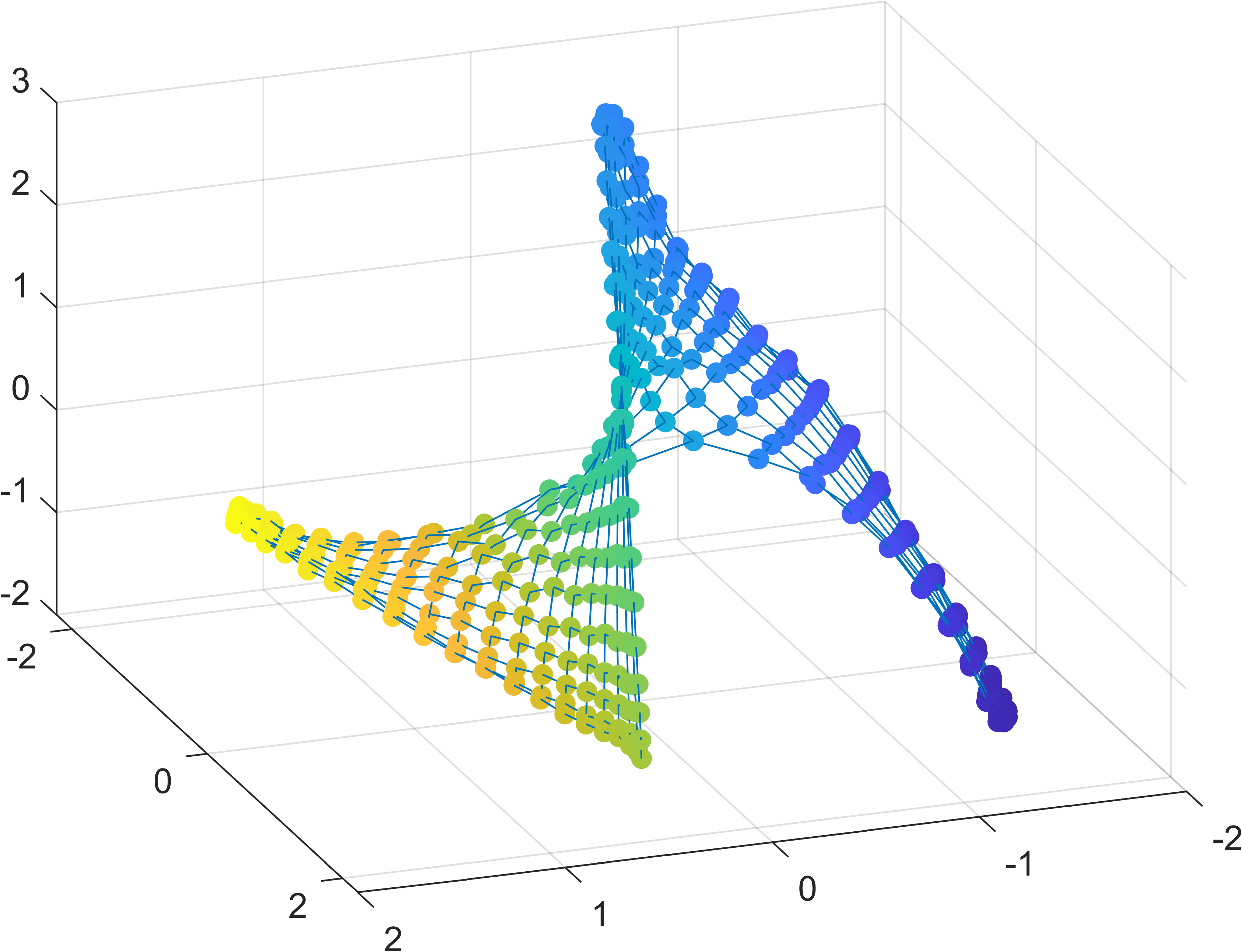}
        \caption{}
        \label{fig:Eigenspace_FV}
    \end{subfigure}
    \hfill
    \begin{subfigure}{0.5\textwidth}
        \centering
        \includegraphics[width=0.7\textwidth]{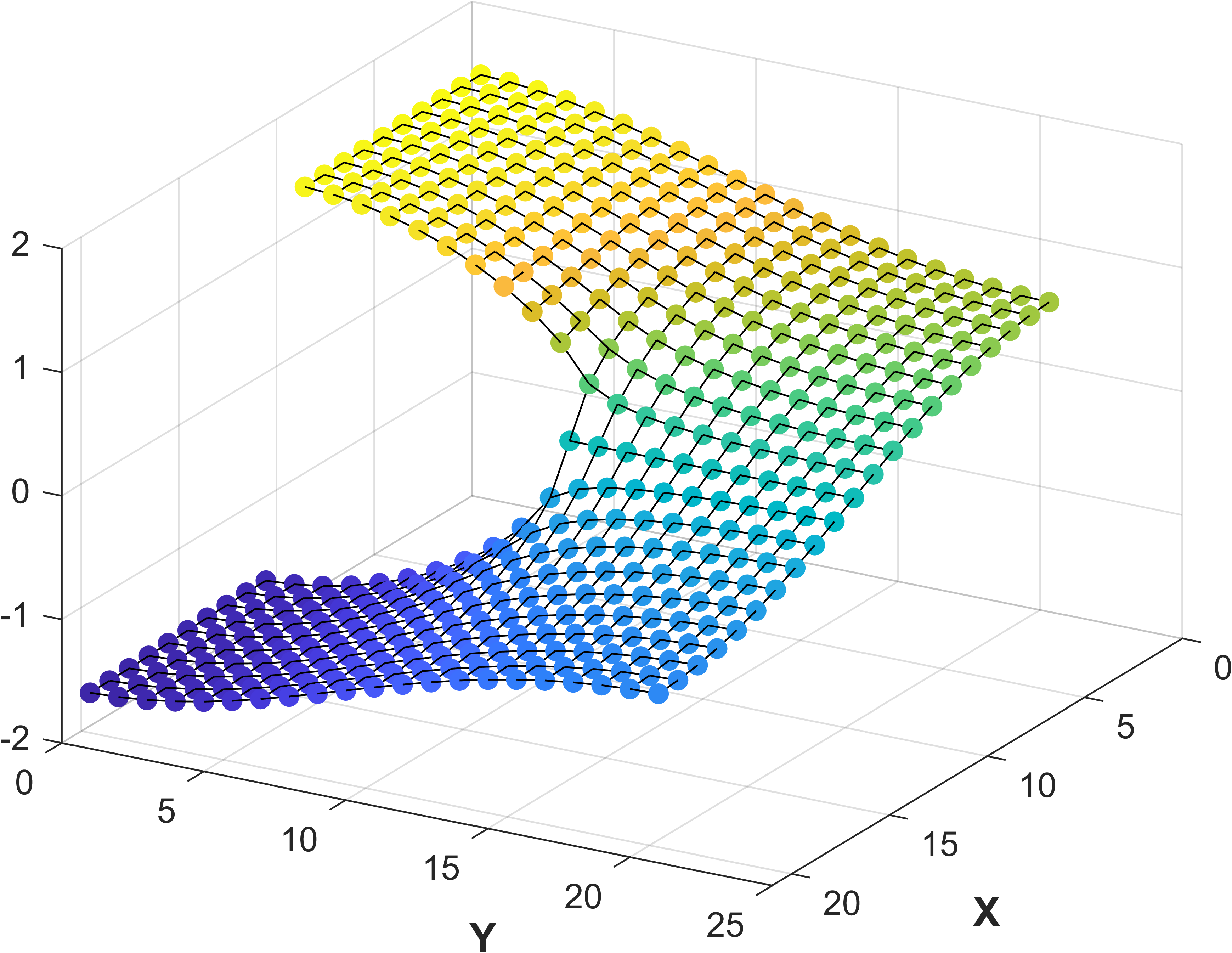}
        \caption{}
        \label{fig:Complex_Geometry_FV}
    \end{subfigure}
    \caption{(a) Visualised graph in the eigenspace, (b) Visualised graph layout in higher-dimensional topological space.}
    \label{fig:whole_FV}
\end{figure}

Fig.\ref{fig:Eigenspace_FV} illustrates the Fiedler vector in the form of graph labels, while Fig.\ref{fig:Complex_Geometry_FV} showcases how it is projected back onto the component. It becomes clear that the Fiedler vector's value can unveil the geometric information of the component, or in other terms, it mirrors the shape of the component. The new input space is then developed as $[\boldsymbol{x},t,z]$ where $z$ is defined as the obtained Fiedler vector. The implementation of such a case usually uses a finite element mesh to construct the graph and get the Fiedler vector value on the graph nodes. Further FE extrapolation is employed to the value in the whole field. An exemplified code for calculating the Fiedler vector from a FE mesh is available at \href{https://github.com/hl4220/Physics-informed-Neural-Network-with-Graph-Embedding.git}{https://github.com/hl4220/Physics-informed-Neural-Network-with-Graph-Embedding.git}.

\section{Results} \label{sec:s3}
This section showcases two case studies: one focuses on a heat conduction model, and the other on crack modelling in solid mechanics. Both problems are examined using traditional Physics-informed Neural Network (PINN) and the enhanced PINN with Graph Embedding (GPINN). The process of graph construction for approximating extra dimensions is explained, and the results are illustrated, with high-precision finite element method solutions serving as a reference point for comparison.

\subsection{Model of heat propagation} \label{sec:s3_1}
A steady-state temperature field follows the Poisson's equation for balance of internal heat sources as well as thermal boundary conditions \cite{cai2021physics,grossmann2023can}:
\begin{equation}
\begin{gathered}
\Delta u(\boldsymbol{x})=f(\boldsymbol{x}), \quad \boldsymbol{x} \in \Omega, \\
u(\boldsymbol{x})=u_{\partial}(\boldsymbol{x}), \quad \boldsymbol{x} \in \partial \Omega, \\
\nabla u(\boldsymbol{x}) \cdot \boldsymbol{n}=v_{\partial}(\boldsymbol{x}) \quad \boldsymbol{x} \in \partial \Omega,
\end{gathered}
\label{e5}
\end{equation}
where the second the third equations denote the Dirichlet and Neumann boundary conditions, respectively, $\Delta$ denotes the Laplace operator. The loss function employed in PINN is expressed as:
\begin{equation}
\begin{gathered}
\mathcal{L} = \omega_1 \mathcal{L}_{\mathrm{PDE}}+\omega_2 \mathcal{L}_{\mathrm{Data}} +\omega_4 \mathcal{L}_{\mathrm{BC}}, \\
\mathcal{L}_{\mathrm{PDE}} = \frac{1}{N_p} \sum_{i=1}^{N_p}\left|\Delta u(\boldsymbol{x}_p) - f(\boldsymbol{x}_p)\right|^2, \\
\mathcal{L}_{\mathrm{data}} = \frac{1}{N_D} \sum_{i=1}^{N_D}\left|u(\boldsymbol{x}_D) - u^*(\boldsymbol{x}_D)\right|^2, \\
\mathcal{L}_{\mathrm{BC}} =
\frac{1}{N_{dbc}} \sum_{i=1}^{N_{dbc}}\left|u(\boldsymbol{x}_{dbc}) - u_{\partial}(\boldsymbol{x}_{dbc})\right|^2 + \\
\frac{1}{N_{nbc}} \sum_{i=1}^{N_{nbc}}\left|\nabla u(\boldsymbol{x}_{nbc}) \cdot \boldsymbol{n} - v_{\partial}(\boldsymbol{x}_{nbc})\right|^2 ,
\label{e6}
\end{gathered}
\end{equation}

\begin{figure}
\centering
\includegraphics[width=0.36\textwidth]{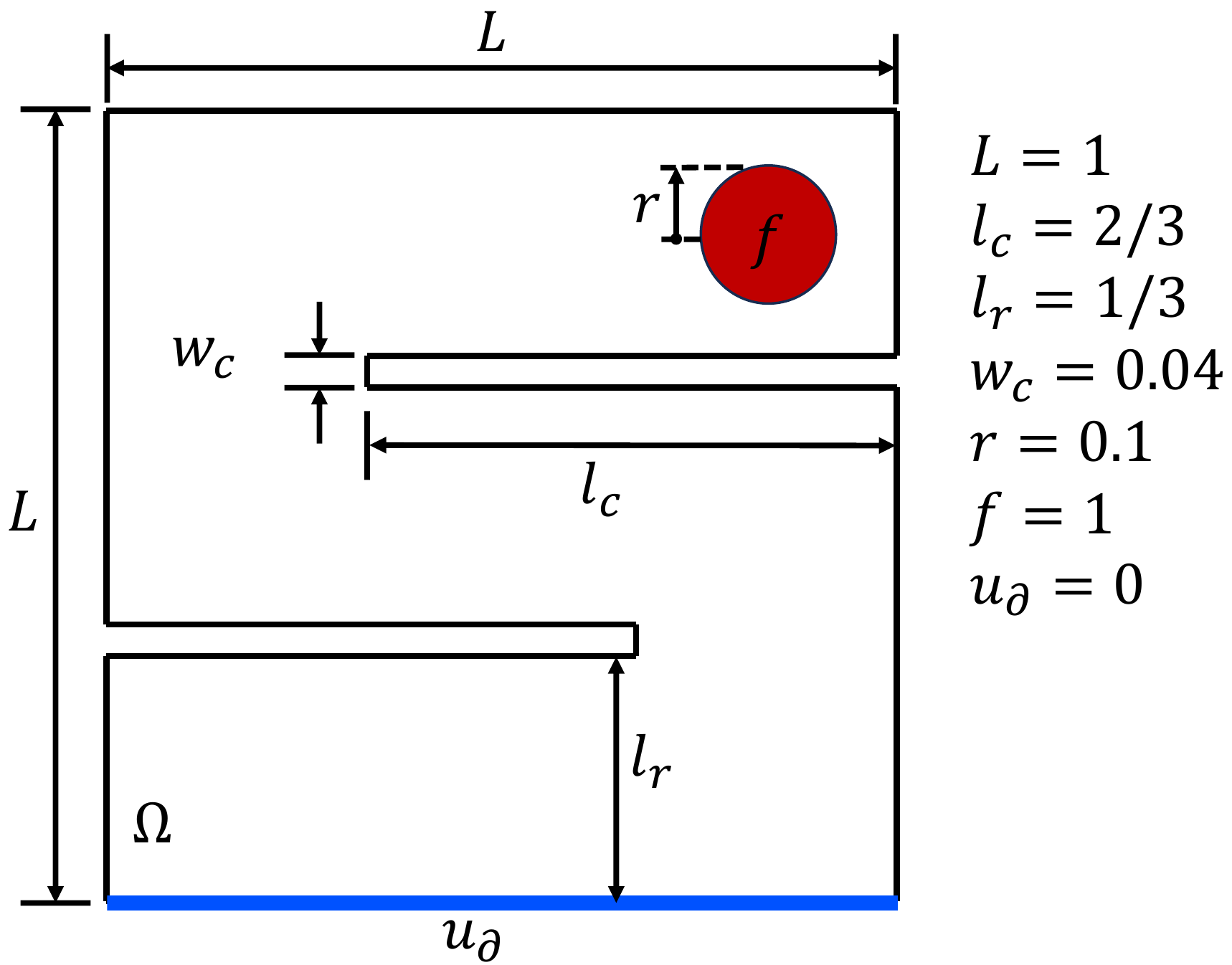}
\caption{Schematic of the heat propagation problem. The domain of the 2D 'house' is defined as $\Omega$; there is a heat source $f$ in $\Omega$ ; the Dirichlet boundary conditoin is assigned on the boundary at the bottom of $\Omega$ that represents a 'window' whose temperature $u_{\partial}$ is the same as 'outside'.}
\label{Fig_3}
\end{figure}

In this work, the heat propagation problem is defined to find the steady temperature field in a 2D 'house' with the domain and boundary conditions demonstrated in Fig.\ref{Fig_3}. The house presents a square area with two walls that separate the house in some extent. A circular heat source (highlighted in red in Fig.\ref{Fig_3}) is located at a corner of the house and the boundary away from the heat source is the house's 'window' that is so thin that keeps a temperature identical to 'outside'.

\begin{figure*}
\centering
\includegraphics[width=0.98\textwidth]{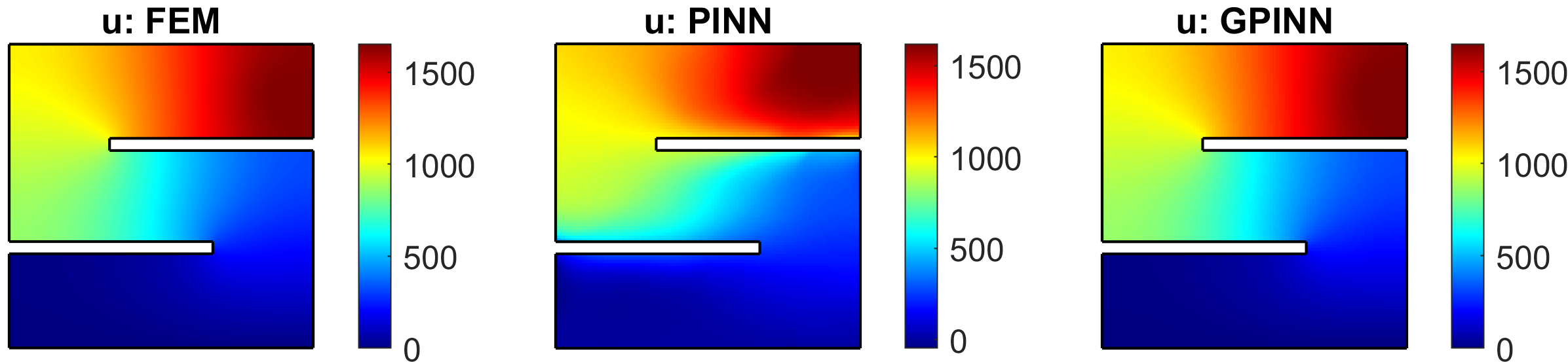}
\caption{Reference(FEM) and Sample(NN) solutions of the steady temperature field.}
\label{Fig_4}
\end{figure*}

\begin{figure*}
\centering
\includegraphics[width=0.57\textwidth]{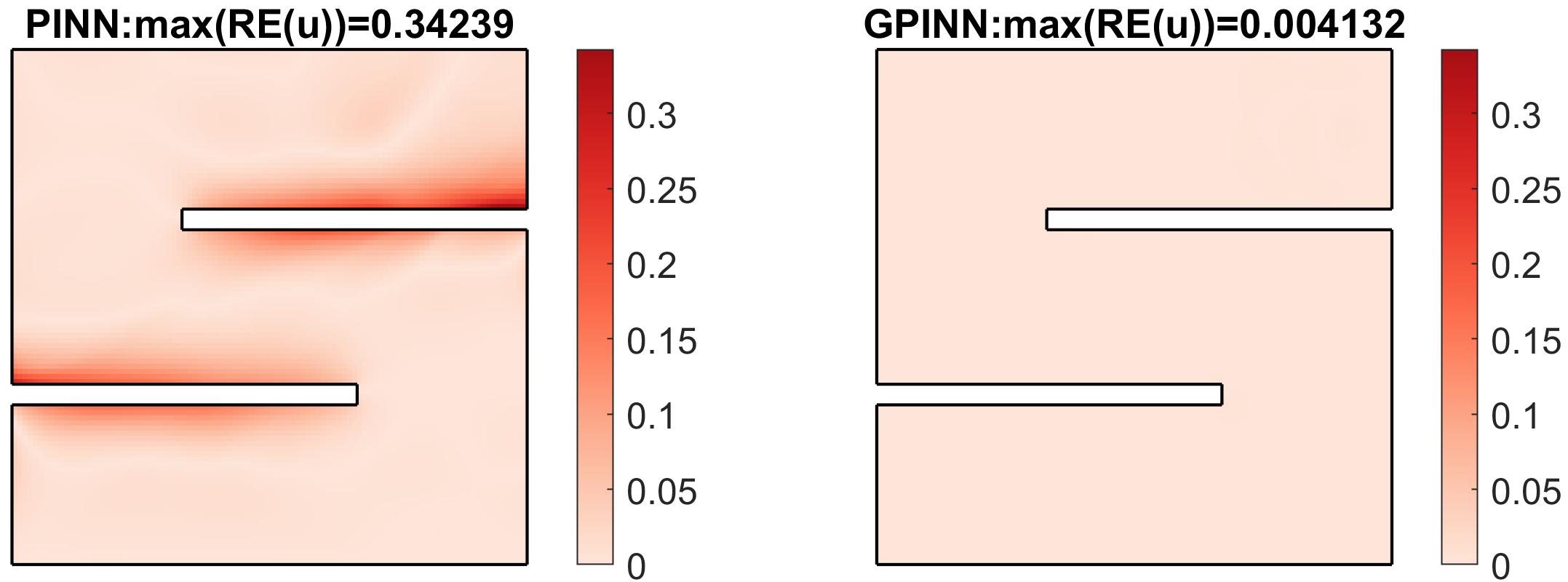}
\caption{Relative errors ($RE$) of the NN solutions to the reference FEM solution: $\operatorname{RE}(u)=\left|u-u^*\right| / \max \left(\left|u^*\right|\right)$. The subfigure on the left side is the relative error of PINN while the right one represents the GPINN.}
\label{Fig_5}
\end{figure*}

\begin{figure}
\centering
\includegraphics[width=0.40\textwidth]{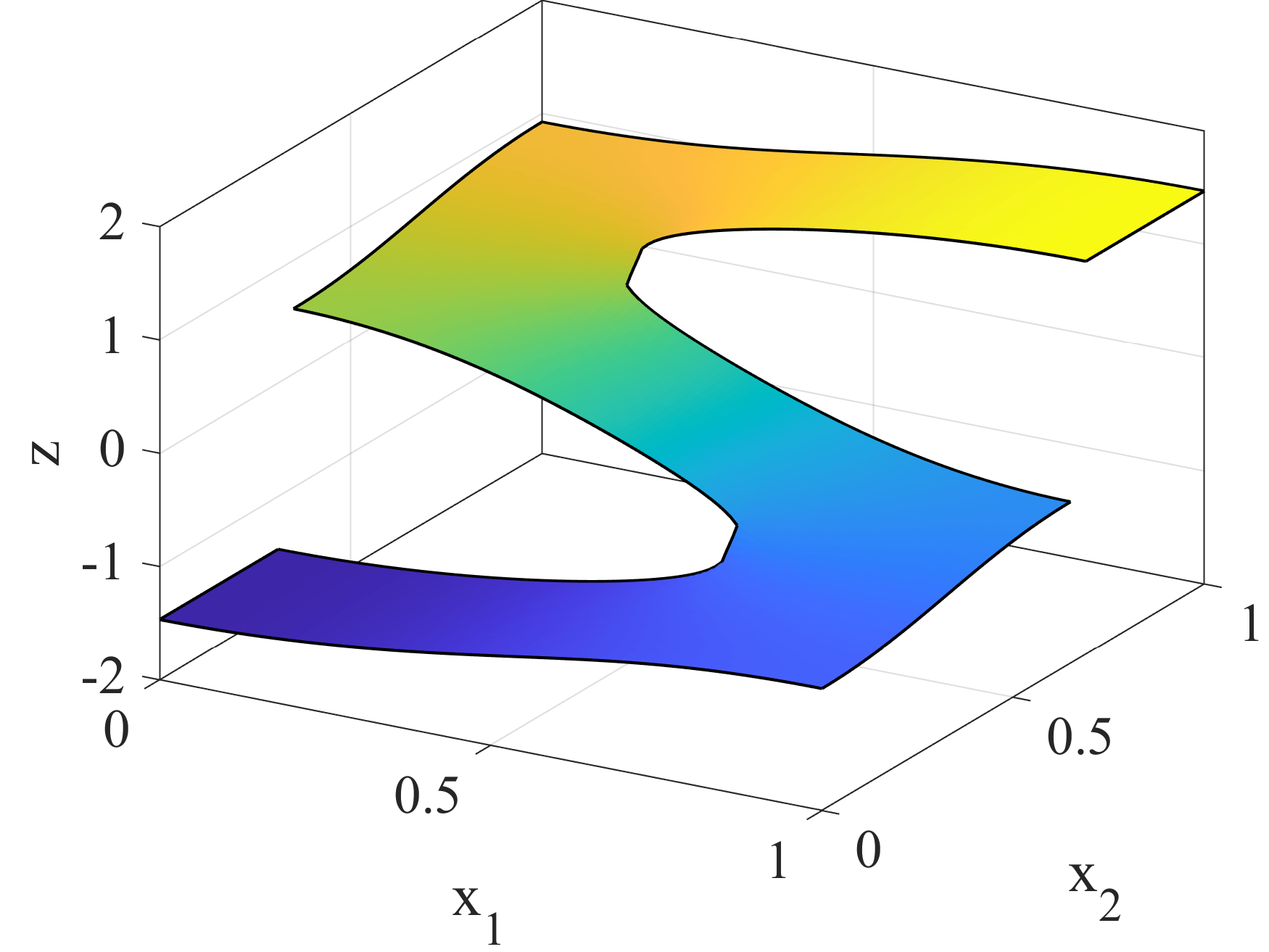}
\caption{Input spatial model in the topological space. A 2D input space is expanded to 2D space by the incorporated extra dimensions. The extra dimensions is determined by the Fiedler vector to keep physical consistency of the input model. The colour map represents the value distribution of the extra-dimension.}
\label{Fig_6}
\end{figure}

The exemplified problem is investigated by both PINN and GPINN. The results are compared to the FEM results derived from a very fine mesh ($256 \times 256$) in Fig.\ref{Fig_4}. Fig.\ref{Fig_5} presents the relative errors of NN to the FEM results. From Fig.\ref{Fig_4} and \ref{Fig_5}, GPINN produces satisfactory outcomes when compared to the reference FEM results, particularly in this problem where the traditional PINN exhibits subpar performance.
Fig.\ref{Fig_4}(b) shows the limitation of traditional PINN in dealing with relatively discontinuous field. This reveals the drawback of PINN which employed the Euclidean distance directly in modelling which does not really correspond to the real physical distance. After incorporating an extra-dimension that expand the problem from 2D Euclidean space to 3D topological space, the PINN model is highly enhanced. The extra-dimension $z$ is determined by the Fiedler vector in graph theory as introduced in section.\nameref{sec:s2_4}. The input 3D coordinate model in the topological space is shown in Fig.\ref{Fig_6}.

\subsection{Model of single-side crack} \label{sec:s3_2}

\begin{figure}[h!]
\centering
\includegraphics[width=0.30\textwidth]{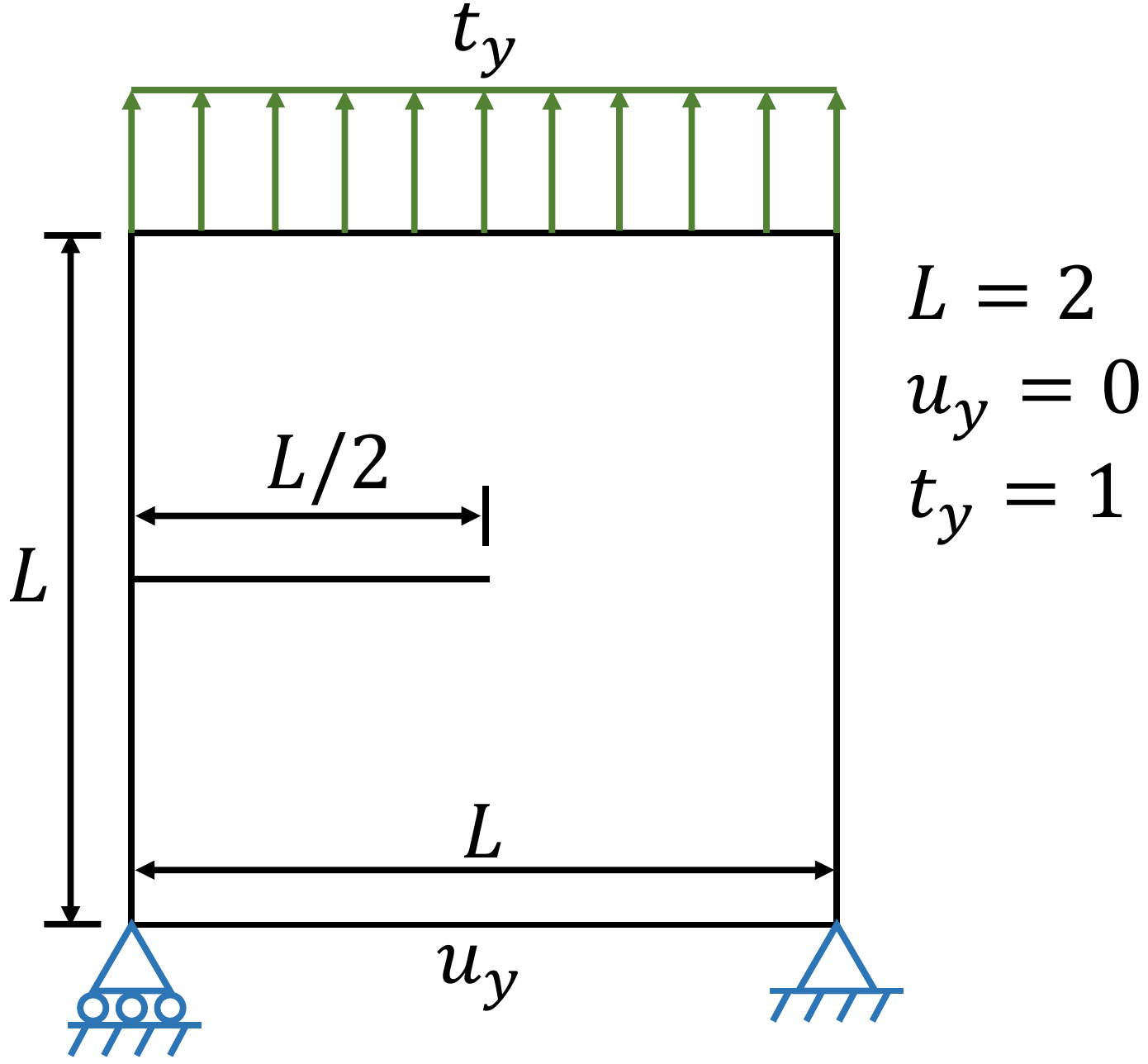}
\caption{Schematic of the single-side crack tensile test. The Dirichlet and Neumman boundary conditions are assigned as indicated.}
\label{Fig_7}
\end{figure}
\begin{figure*}[h!]
\centering
\includegraphics[width=0.95\textwidth]{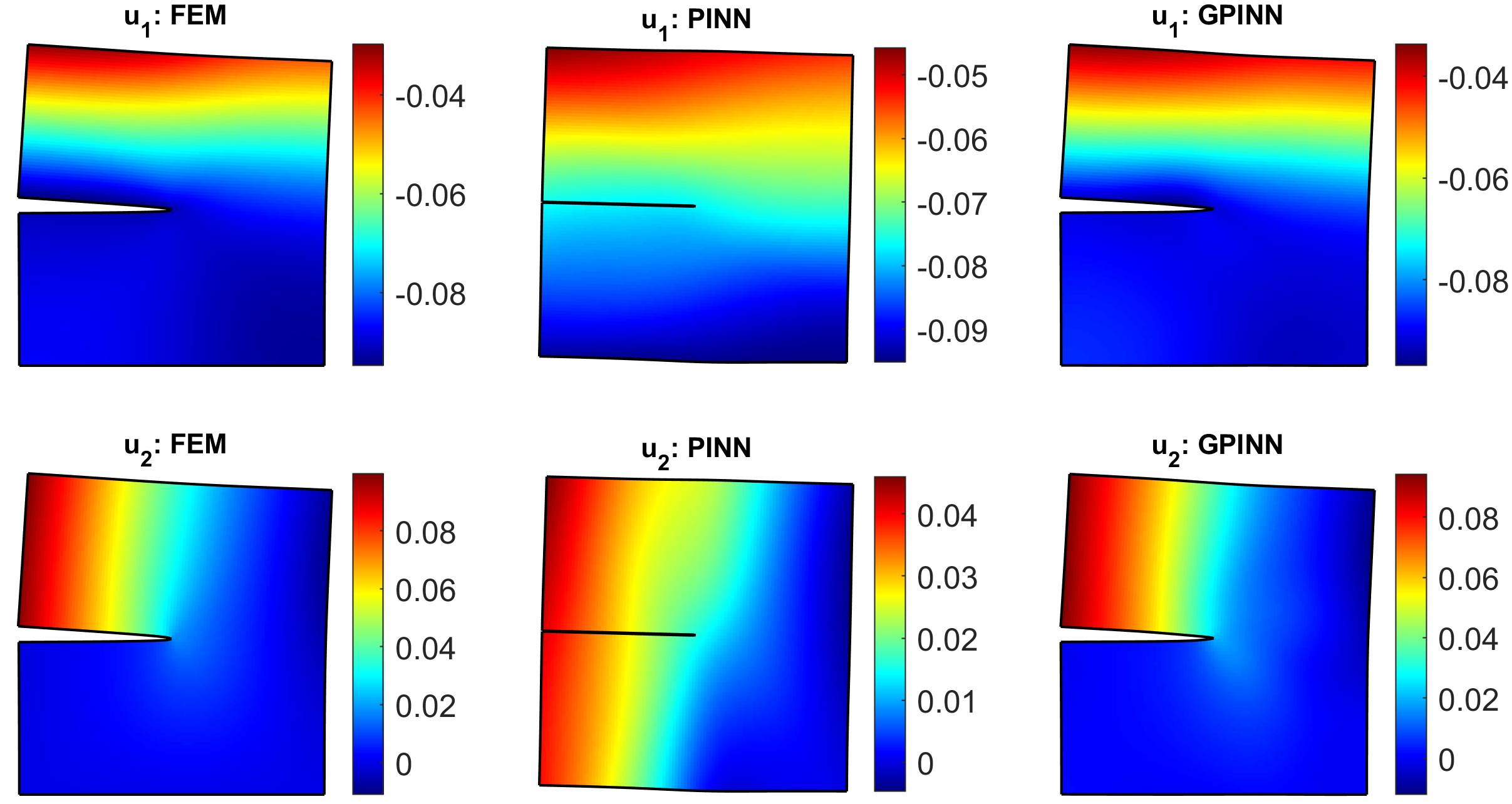}
\caption{Reference (FEM) and Sample (NN) solutions of the steady temperature field.}
\label{Fig_8}
\end{figure*}
\begin{figure*}[h!]
\centering
\includegraphics[width=0.60\textwidth]{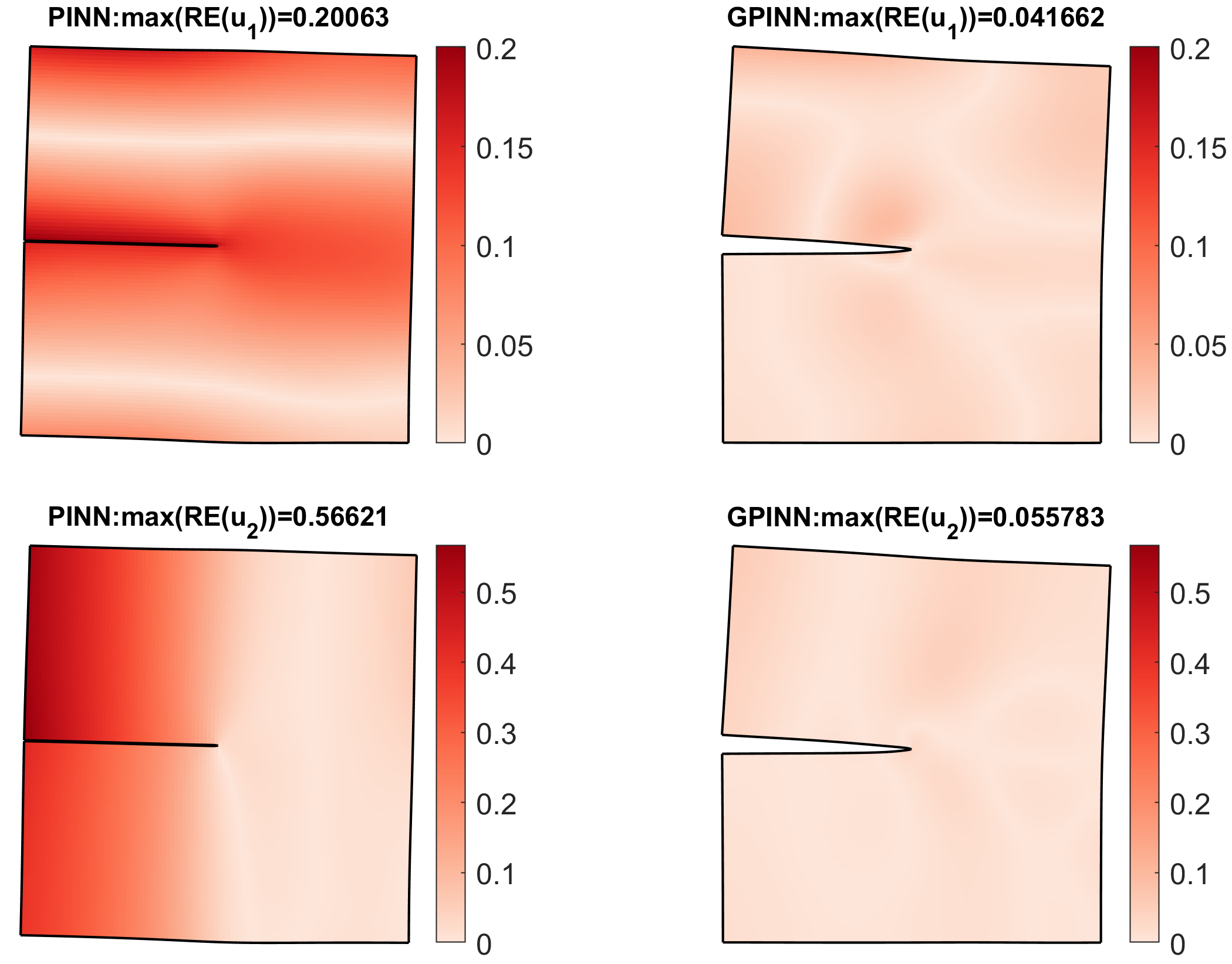}
\caption{Relative errors ($RE$) of the NN solutions to the reference FEM solution: $\operatorname{RE}(u)=\left|u-u^*\right| / \max \left(\left|u^*\right|\right)$. The subfigures on the left side are the relative errors of PINN while the right ones represent the GPINN.}
\label{Fig_9}
\end{figure*}
\begin{figure}[h!]
\centering
\includegraphics[width=0.40\textwidth]{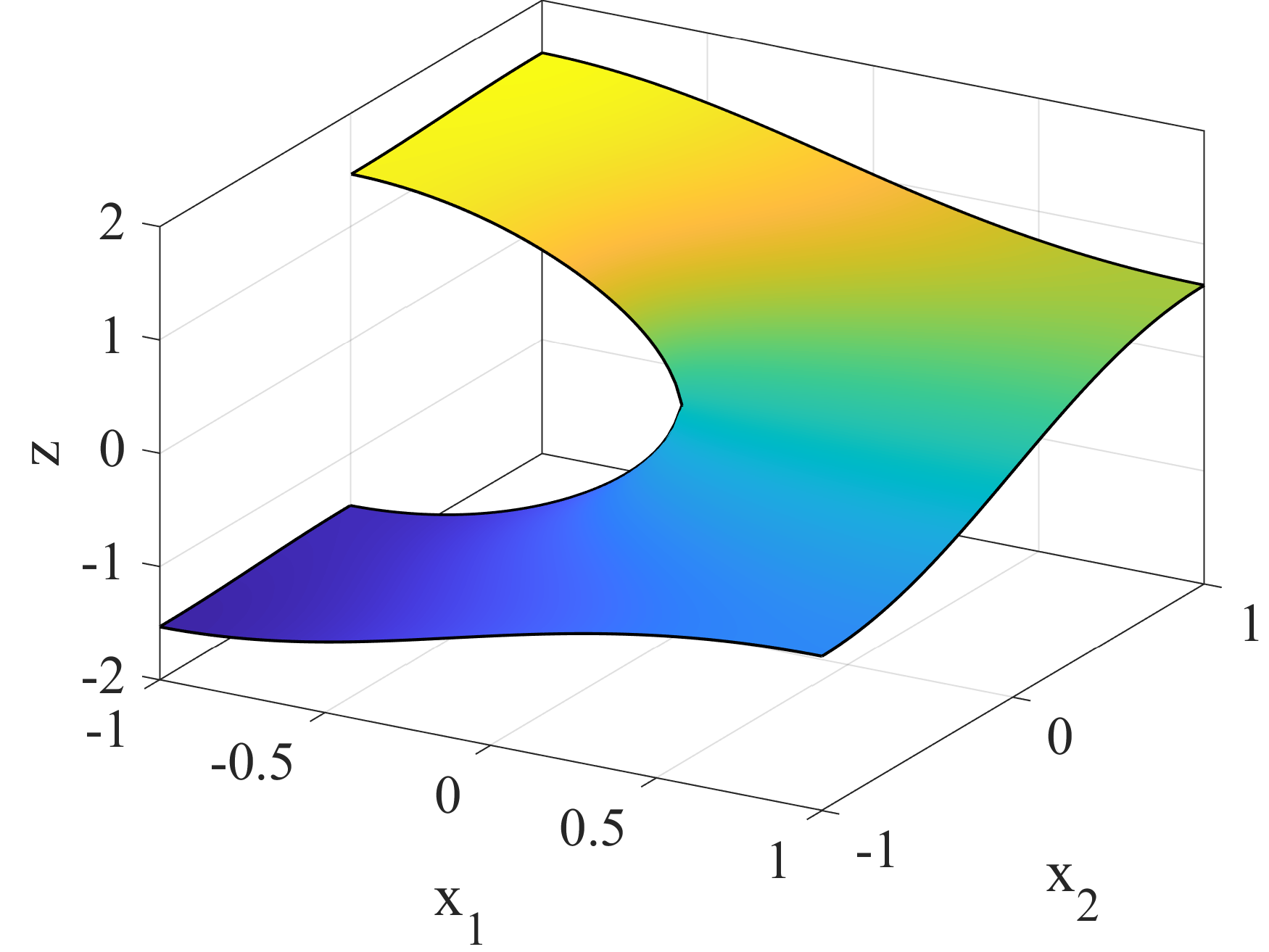}
\caption{Input spatial model in the topological space. A 2D input space is expanded to 2D space by the incorporated extra dimensions. The extra dimensions is determined by the Fiedler vector to keep physical consistency of the input model. The colour map represents the value distribution of the extra-dimension.}
\label{Fig_10}
\end{figure}
The second case conducts a linear elastic simulation of a single-side crack model. Traditional PINN usually shows its weakness in dealing with crack problems due to the strong discontinuity near the crack \cite{zheng2022physics,goswami2020transfer,haghighat2021physics}. The governing equation of the linear elasticity is stated as:
\begin{equation}
\begin{gathered}
\boldsymbol{\nabla} \cdot \boldsymbol{\sigma}(\boldsymbol{x})=\mathbf{0}, \boldsymbol{x} \in \Omega, \\
\boldsymbol{\sigma}(\boldsymbol{x}) \cdot \boldsymbol{n}(\boldsymbol{x})=\boldsymbol{t}(\boldsymbol{x}), \boldsymbol{x} \in \partial \Omega,  \\
\boldsymbol{u}(\boldsymbol{x})=\boldsymbol{u}_{\partial }(\boldsymbol{x}), \boldsymbol{x} \in \partial \Omega, \\
\end{gathered}
\label{e7}
\end{equation}
where $\boldsymbol{u}$ and $\boldsymbol{\sigma}$ denotes the displacement and stress vectors, respectively, being effective on the domain $\Omega$ with the boundary
$\partial \Omega$. The second and third equations in Eq.\ref{e7} denote the Neumann and Dirichlet boundary conditions, respectively, in which $\boldsymbol{t}$ and $\boldsymbol{u}_{b c}$ are the applied traction and displacement conditions. Note that the output field $\boldsymbol{u}$ is a vector field that differs from the temperature field targeting at in the last problem \cite{arora2022physics}. The bold symbols utilised in those equations thus refer to vectors rather than scalers as shown Eq.\ref{e7}.

In this work, the energy-based loss function is employed herein for the PINN model. Energy-based loss function aims at minimising the potential energy of the entire structure, which considers global information and requires less differential order compared to collocation loss function \cite{bai2022introduction}. It has also been investigated to be better in modelling crack problems \cite{zheng2022physics}. The loss function employed in PINN in this problem is stated as \cite{bai2022introduction,zheng2022physics}:
\begin{equation}
\begin{gathered}
\mathcal{L} = \omega_1 \mathcal{L}_{PDE} + \omega_2 \mathcal{L}_{\mathrm{Data}} +\omega_4 \mathcal{L}_{\mathrm{BC}}, \\
\mathcal{L}_{PDE} = \int_{\Omega} \frac{1}{2} \boldsymbol{\sigma}(\boldsymbol{x}_{p}) \boldsymbol{\varepsilon}(\boldsymbol{x}_{p}) \mathrm{d} \Omega - \\
 \int_{\partial \Omega} \boldsymbol{t}(\boldsymbol{x}_{nbc}) \boldsymbol{u}(\boldsymbol{x}_{nbc}) \mathrm{d} \partial \Omega, \\
\mathcal{L}_{\mathrm{Data}} = \frac{1}{N_D} ( \sum_{i=1}^{N_D}\left\|\boldsymbol{u}(\boldsymbol{x}_D) - \boldsymbol{u}^*(\boldsymbol{x}_D)\right\|^2 + \\
\sum_{i=1}^{N_D}\left\|\boldsymbol{\sigma}(\boldsymbol{x}_D) - \boldsymbol{\sigma}^*(\boldsymbol{x}_D)\right\|^2 ), \\
\mathcal{L}_{\mathrm{BC}} =
\frac{1}{N_{dbc}} \sum_{i=1}^{N_{dbc}}\left\|\boldsymbol{u}(\boldsymbol{x}_{dbc}) - \boldsymbol{u}_{bc}\right\|^2 + \\
\frac{1}{N_{nbc}} \sum_{i=1}^{N_{nbc}}\left\|\boldsymbol{\sigma}(\boldsymbol{x}_{nbc}) \cdot \boldsymbol{n}-\boldsymbol{t}(\boldsymbol{x}_{nbc})\right\|^2 ,
\label{e8}
\end{gathered}
\end{equation}
where $\boldsymbol{\sigma}$ and $\boldsymbol{\varepsilon}$ denote the stress and strain:
\begin{equation}
\begin{gathered}
\boldsymbol{\varepsilon}(\boldsymbol{x}) = \nabla \boldsymbol{u}(\boldsymbol{x}), \\
\boldsymbol{\sigma}(\boldsymbol{x}) = \mathbb{C}:\boldsymbol{\varepsilon}(\boldsymbol{x}), \\
\end{gathered}
\end{equation}
This case study involves a tensile test for a model with a single-sided crack. The problem schematic is depicted in Fig.\ref{Fig_7}. The results of both PINN and GPINN methods are presented in Fig.\ref{Fig_8}, alongside the reference FEM results. The distribution of relative error is displayed in Fig.\ref{Fig_9}. From these illustrations, it is evident that the GPINN method yields promising results in comparison to the reference FEM results. Traditional PINN, however, seems to fall short in accurately capturing the solution features for such a problem. The input model enriched with extra dimensions is presented in Fig.\ref{Fig_10}.

\section{Conclusion} \label{sec:s4}
In this work, we present a noval method to perform PINN in the topological, i.e. graph space for better capturing the physical characteristic of a structure. It is achieved by incorporating extra dimensions into the input space, creating a graph-based spatial model which offers a better capture of the pathological property of a structure. The extra dimensions are derived from the graph theory, utilizing the Fiedler vector, which serves as an approximate optimal solution for the graph space layout. Our results illustrate that the graph embedding significantly enhances PINN, particularly for dealing with problems within complex domains.

This method essentially transforms PINN modelling from a Euclidean space into a graph-consistent space. Its potential in engineering applications is significant, considering its ease of implementation and the substantial improvements it brings to PINN's performance. Moreover, the method proposes a possible solution for addressing fracture and crack problems that involve highly discontinuous fields.

In the work presented here, only one extra dimension, the Fiedler vector, was incorporated into our modelling approach. Future research will focus on integrating more extra dimensions that can provide valuable insights for the modelling and solving process in PINN.

\bibliographystyle{aaai}
\bibliography{References}

\begin{thebibliography}{}

\bibitem[\protect\citeauthoryear{Arora \bgroup et al\mbox.\egroup
  }{2022}]{arora2022physics}
Arora, R.; Kakkar, P.; Dey, B.; and Chakraborty, A.
\newblock 2022.
\newblock Physics-informed neural networks for modeling rate-and
  temperature-dependent plasticity.
\newblock {\em arXiv preprint arXiv:2201.08363}.

\bibitem[\protect\citeauthoryear{Bai \bgroup et al\mbox.\egroup
  }{2022}]{bai2022introduction}
Bai, J.; Jeong, H.; Batuwatta-Gamage, C.; Xiao, S.; Wang, Q.; Rathnayaka, C.;
  Alzubaidi, L.; Liu, G.-R.; and Gu, Y.
\newblock 2022.
\newblock An introduction to programming physics-informed neural network-based
  computational solid mechanics.
\newblock {\em arXiv preprint arXiv:2210.09060}.

\bibitem[\protect\citeauthoryear{Banuelos and Burdzy}{1999}]{banuelos1999hot}
Banuelos, R., and Burdzy, K.
\newblock 1999.
\newblock On the hot spots conjecture of j rauch.
\newblock {\em Journal of functional analysis} 164(1):1--33.

\bibitem[\protect\citeauthoryear{Cai \bgroup et al\mbox.\egroup
  }{2021}]{cai2021physics}
Cai, S.; Wang, Z.; Wang, S.; Perdikaris, P.; and Karniadakis, G.~E.
\newblock 2021.
\newblock Physics-informed neural networks for heat transfer problems.
\newblock {\em Journal of Heat Transfer} 143(6).

\bibitem[\protect\citeauthoryear{Chen \bgroup et al\mbox.\egroup
  }{2020}]{chen2020physics}
Chen, Y.; Lu, L.; Karniadakis, G.~E.; and Dal~Negro, L.
\newblock 2020.
\newblock Physics-informed neural networks for inverse problems in nano-optics
  and metamaterials.
\newblock {\em Optics express} 28(8):11618--11633.

\bibitem[\protect\citeauthoryear{Chung \bgroup et al\mbox.\egroup
  }{2011}]{chung2011hot}
Chung, M.~K.; Seo, S.; Adluru, N.; and Vorperian, H.~K.
\newblock 2011.
\newblock Hot spots conjecture and its application to modeling tubular
  structures.
\newblock In {\em Machine Learning in Medical Imaging: Second International
  Workshop, MLMI 2011, Held in Conjunction with MICCAI 2011, Toronto, Canada,
  September 18, 2011. Proceedings 2},  225--232.
\newblock Springer.

\bibitem[\protect\citeauthoryear{Goswami \bgroup et al\mbox.\egroup
  }{2020}]{goswami2020transfer}
Goswami, S.; Anitescu, C.; Chakraborty, S.; and Rabczuk, T.
\newblock 2020.
\newblock Transfer learning enhanced physics informed neural network for
  phase-field modeling of fracture.
\newblock {\em Theoretical and Applied Fracture Mechanics} 106:102447.

\bibitem[\protect\citeauthoryear{Grossmann \bgroup et al\mbox.\egroup
  }{2023}]{grossmann2023can}
Grossmann, T.~G.; Komorowska, U.~J.; Latz, J.; and Sch{\"o}nlieb, C.-B.
\newblock 2023.
\newblock Can physics-informed neural networks beat the finite element method?
\newblock {\em arXiv preprint arXiv:2302.04107}.

\bibitem[\protect\citeauthoryear{Haghighat \bgroup et al\mbox.\egroup
  }{2021}]{haghighat2021physics}
Haghighat, E.; Raissi, M.; Moure, A.; Gomez, H.; and Juanes, R.
\newblock 2021.
\newblock A physics-informed deep learning framework for inversion and
  surrogate modeling in solid mechanics.
\newblock {\em Computer Methods in Applied Mechanics and Engineering}
  379:113741.

\bibitem[\protect\citeauthoryear{Raissi, Perdikaris, and
  Karniadakis}{2019}]{raissi2019physics}
Raissi, M.; Perdikaris, P.; and Karniadakis, G.~E.
\newblock 2019.
\newblock Physics-informed neural networks: A deep learning framework for
  solving forward and inverse problems involving nonlinear partial differential
  equations.
\newblock {\em Journal of Computational physics} 378:686--707.

\bibitem[\protect\citeauthoryear{Zheng \bgroup et al\mbox.\egroup
  }{2022}]{zheng2022physics}
Zheng, B.; Li, T.; Qi, H.; Gao, L.; Liu, X.; and Yuan, L.
\newblock 2022.
\newblock Physics-informed machine learning model for computational fracture of
  quasi-brittle materials without labelled data.
\newblock {\em International Journal of Mechanical Sciences} 223:107282.

\bibitem[\protect\citeauthoryear{Zienkiewicz, Taylor, and
  Taylor}{2000}]{zienkiewicz2000finite}
Zienkiewicz, O.~C.; Taylor, R.~L.; and Taylor, R.~L.
\newblock 2000.
\newblock {\em The finite element method: solid mechanics}, volume~2.
\newblock Butterworth-heinemann.

\end{thebibliography}

\end{document}